\PassOptionsToPackage{unicode}{hyperref}
\PassOptionsToPackage{hyphens}{url}
\PassOptionsToPackage{dvipsnames,svgnames,x11names}{xcolor}
\documentclass[
]{article}
\usepackage{amsmath,amssymb}
\usepackage{lmodern}
\usepackage{iftex}
\ifPDFTeX
  \usepackage[T1]{fontenc}
  \usepackage[utf8]{inputenc}
  \usepackage{textcomp} 
\else 
  \usepackage{unicode-math}
  \defaultfontfeatures{Scale=MatchLowercase}
  \defaultfontfeatures[\rmfamily]{Ligatures=TeX,Scale=1}
\fi
\IfFileExists{upquote.sty}{\usepackage{upquote}}{}
\IfFileExists{microtype.sty}{
  \usepackage[]{microtype}
  \UseMicrotypeSet[protrusion]{basicmath} 
}{}
\makeatletter
\@ifundefined{KOMAClassName}{
  \IfFileExists{parskip.sty}{%
    \usepackage{parskip}
  }{
    \setlength{\parindent}{0pt}
    \setlength{\parskip}{6pt plus 2pt minus 1pt}}
}{
  \KOMAoptions{parskip=half}}
\makeatother
\usepackage{xcolor}
\usepackage{graphicx}
\makeatletter
\def\maxwidth{\ifdim\Gin@nat@width>\linewidth\linewidth\else\Gin@nat@width\fi}
\def\maxheight{\ifdim\Gin@nat@height>\textheight\textheight\else\Gin@nat@height\fi}
\makeatother
\setkeys{Gin}{width=\maxwidth,height=\maxheight,keepaspectratio}
\makeatletter
\def\fps@figure{htbp}
\makeatother
\newlength{\cslhangindent}
\newlength{\csllabelwidth}
\newlength{\cslentryspacingunit} 
 {
  \setlength{\parindent}{0pt}
  \ifodd #1
  \let\oldpar\par
  \def\par{\hangindent=\cslhangindent\oldpar}
  \fi
  \setlength{\parskip}{#2\cslentryspacingunit}
 }%
 {}
\usepackage{calc}

\ifLuaTeX
\usepackage[bidi=basic]{babel}
\else
\usepackage[bidi=default]{babel}
\fi
\babelprovide[main,import]{american}

\def\languageshorthands#1{}
\ifLuaTeX
  \usepackage{selnolig}  
\fi
\IfFileExists{bookmark.sty}{\usepackage{bookmark}}{\usepackage{hyperref}}
\IfFileExists{xurl.sty}{\usepackage{xurl}}{} 
\hypersetup{
  pdftitle={PyBADS: Fast and robust black-box optimization in Python},
  pdfauthor={Gurjeet Sangra Singh, Luigi Acerbi},
  pdflang={en-US},
  colorlinks=true,
  linkcolor={Maroon},
  filecolor={Maroon},
  citecolor={Blue},
  urlcolor={Blue},
  pdfcreator={LaTeX via pandoc}}

\title{PyBADS: Fast and robust black-box optimization in Python}

\usepackage[affil-it]{authblk}
\author[1, 3%
  ]{Gurjeet Sangra Singh%
    \,%
    }
\author[2%
  ]{Luigi Acerbi%
    \,%
    }

\affil[1]{University of Geneva}
\affil[2]{University of Helsinki}
\affil[3]{University of Applied Sciences and Art Western Switzerland (HES-SO)}

\date{11 June 2023}

\usepackage[backend=biber, style=authoryear-comp,
sorting=ynt, sortcites=true,
maxnames=99, maxcitenames=2,
uniquename=full,uniquelist=false, dashed=false
]{biblatex}
\usepackage{csquotes}

\begin{document}
\maketitle

\hypertarget{summary}{%
\section{Summary}\label{summary}}
PyBADS is a Python implementation of the Bayesian Adaptive Direct Search (BADS) algorithm for fast and robust \textit{black-box} optimization \parencite{acerbi2017practical}. BADS is an optimization algorithm designed to efficiently solve difficult optimization problems where the objective function is rough (non-convex, non-smooth), mildly expensive (e.g., the function evaluation requires more than 0.1 seconds), possibly noisy, and gradient information is unavailable. With BADS, these issues are well addressed, making it an excellent choice for fitting computational models using methods such as maximum-likelihood estimation.
The algorithm scales efficiently to black-box functions with up to $D \approx 20$ continuous input parameters and supports bounds or no constraints. PyBADS comes along with an easy-to-use Pythonic interface for running the algorithm and inspecting its results. PyBADS only requires the user to provide a Python function for evaluating the target function, and optionally other constraints.

Extensive benchmarks on both artificial test problems and large real model-fitting problems models drawn from cognitive, behavioral and computational neuroscience, show that BADS performs on par with or better than many other common and state-of-the-art optimizers \parencite{acerbi2017practical}, making it a general model-fitting tool which provides fast and robust solutions. 

\hypertarget{statement-of-need}{%
\section{Statement of need}\label{statement-of-need}}

Many optimization problems in science and engineering involve complex and expensive simulations or numerical approximations such that the target function can only be evaluated at a point with moderate to high cost, possibly yielding stochastic outcomes, and gradients are unavailable (or exceedingly expensive) -- the typical \textit{black-box} scenario. There is a large landscape of derivative-free optimization algorithms for tackling black-box problems  \parencite{Rios2013}, many of which follow variants of direct-search methods \parencite{MADS, stoMADS, orthoMADS, deng2006adaptation}.
Despite their theoretical guarantees, direct search methods require a large number of function evaluations and have limited support for handling stochastic targets. 

Conversely, Bayesian Optimization (BayesOpt) is a recently popular family of methods that has shown effectiveness in solving \textit{very} costly black-box problems in machine learning and engineering with very few, possibly noisy, function evaluations \parencite{garnett_bayesoptbook_2023, reviewBO, agnihotri2020exploring}. However, BayesOpt requires specific technical knowledge to be implemented or tuned beyond simple tasks, since vanilla BayesOpt applied to complex real-world problems can be strongly affected by deviations from the algorithm's assumptions (model misspecification), a problem rarely dealt with in current implementations. Moreover, traditional BayesOpt methods assume \textit{highly expensive} target functions (e.g., with evaluation costs of hours or more), whereas many computational models might only have a \textit{moderate} evaluation cost (e.g., from a fraction of a second to a few seconds), meaning that the optimization algorithm should add only a relatively small overhead. 

PyBADS addresses all these problems as a fast hybrid algorithm that combines the strengths of BayesOpt and the Mesh Adaptive Direct Search \parencite{MADS} method. In contrast to other black-box optimization algorithms, PyBADS is both \textit{fast} in terms of wall-clock time and \textit{sample-efficient} in terms of the number of target evaluations (typically of the order of a few hundred), with support for noisy targets. Moreover, PyBADS does not require any specific tuning and runs off-the-shelf with its well-modularized Python API.

\begin{figure}[th!]
	\centering
	\includegraphics[width=\textwidth]{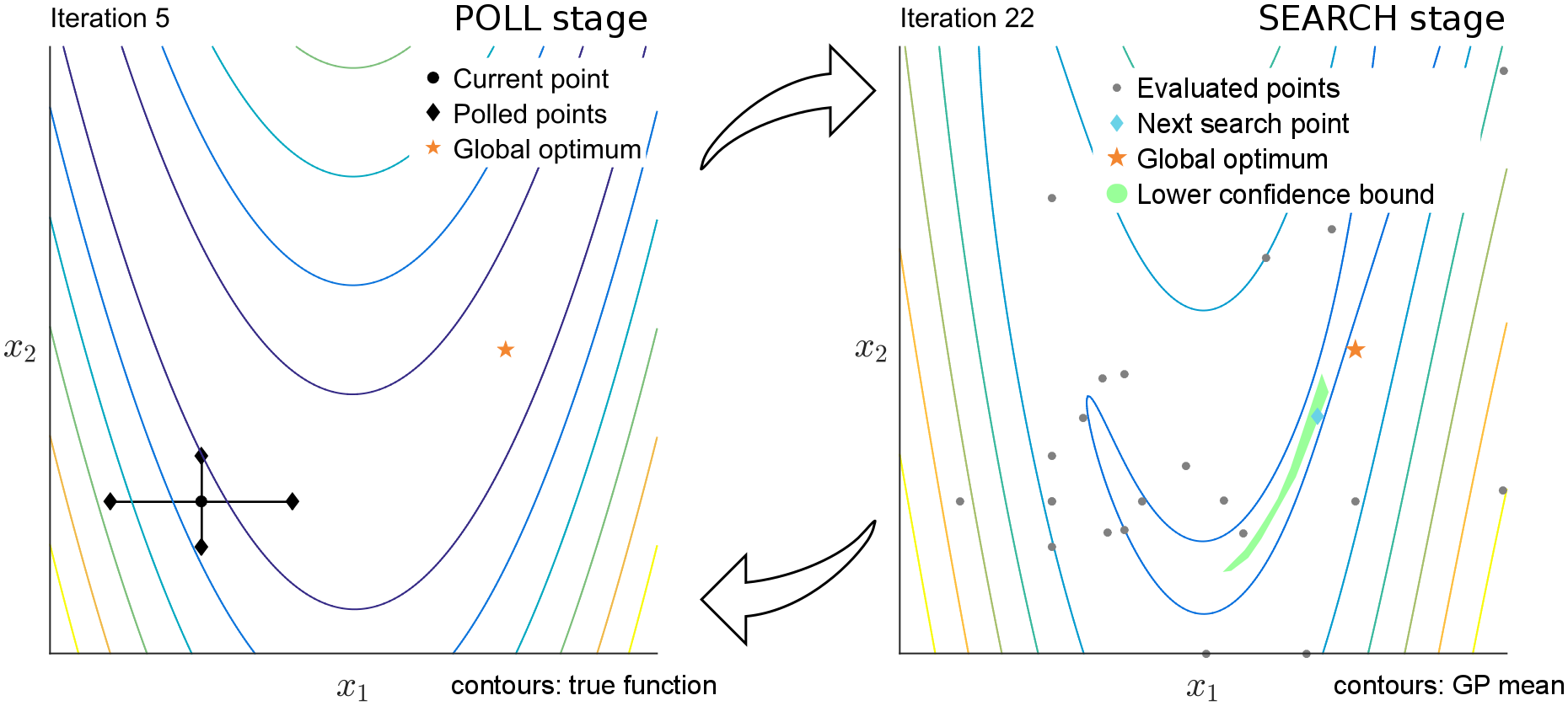}
	\caption{Contour plots and PyBADS exploration of a two-dimensional \href{https://en.wikipedia.org/wiki/Rosenbrock\_function}{Rosenbrock} function. Lines represent the contours of the true target (left) and of the GP surrogate model built during the search stage (right). The solid black diamonds indicate new points chosen by the poll method (here for simplicity a simple orthogonal poll), the grey circles represent the previously sampled points, and the orange solid star represents the global minimum of the function. When switching to the search stage, the blue diamond describes the point selected by the active sampling method based on the lower confidence bound obtained from the GP surrogate model (green region)}
	\label{fig:example}
\end{figure}

\hypertarget{method}{%
\subsection{Method}\label{method}}
PyBADS follows the Mesh Adaptive Direct Search (MADS; \cite{MADS}) schema for minimizing the given objective function. The algorithm alternates between a series of fast local Bayesian Optimization (BayesOpt) steps, referred to as \textit{search stage}, and systematic exploration of the mesh space in a neighborhood of the current point, known as \textit{poll stage}, based on the MADS poll method \parencite{MADS}; see Figure \ref{fig:example}.
Briefly:
\begin{itemize}
    \item In the poll stage, points are evaluated on a mesh by taking steps in one (non-orthogonal) direction at a time, until an improvement is found or all directions have been tried. The step size is doubled in case of success, halved otherwise.
    \item In the search stage, a Gaussian process (GP) surrogate model \parencite{rasmussen_gaussian_2006} of the target function is fit to a local subset of the points evaluated so far. New points to evaluate are quickly chosen according to a \textit{lower confidence bound} strategy that trades off between exploration of uncertain regions (high GP uncertainty) and exploitation of promising solutions (low GP mean). The search switches back to the poll stage after repeated failures to find an improvement over the current point.
\end{itemize}

This alternation between the two stages makes BADS uniquely robust and effective. The poll stage follows a slow but steady "model-free" optimization with theoretical guarantees. Conversely, the search stage exploits a powerful "model-based" GP surrogate to propose potentially large steps, which can be extremely effective if the surrogate is able to approximate the target well. Notably, this strategy is fail-safe in that if the GP fails to locally model the target, the search will fail and PyBADS will fall back to the safer poll method. The points acquired during the poll will afford a construction of a better surrogate at the next search, and so on.
In addition, when the target is noisy, BADS follows some effective heuristics for calibrating the surrogate model, checking the reliability of the predictions, and reassessing the estimated value of the current point in light of the new points. 

Thanks to these techniques, our algorithm has demonstrated high robustness and effectiveness in solving optimization problems with noisy and complex non-convex objective functions.

\hypertarget{related-work}{%
\subsection{Related work}\label{related-work}}

Similarly to PyBADS, relevant libraries have been developed over the years in the area of Bayesian Optimization, such as BoTorch \parencite{balandat2020botorch},  GPflowOpt \parencite{GPflowOpt2017}, Spearmint \parencite{pmlr-v32-snoek14}, among others. Instead, NOMAD \parencite{nomad4paper} is the main reference library for \textit{pattern search} algorithms, and it implements several variants of MADS in C++, by providing Python and Julia interface bindings.

Differently to these algorithms, PyBADS comes with a unique hybrid, fast and robust combination of direct search (MADS) and Bayesian Optimization. This combination of strategies protects against failures of the GP surrogate models – whereas vanilla BayesOpt does not have such fail-safe mechanisms, and can be strongly affected by misspecification of the surrogate GP model. PyBADS has also been designed to avoid problem-specific tuning, making it a generic tool for model fitting. Compared to other approaches, PyBADS also has the advantage of natively accommodating  target functions with heteroskedastic (input-dependent) observation noise. The results of our approach demonstrate that a hybrid Bayesian approach to optimization can be beneficial beyond the domain of costly black-box functions.
Finally, unlike most other BayesOpt packages, targeted to an audience of machine learning researchers, PyBADS comes with a neat API library and well-structured, user-friendly documentation.

PyBADS was developed in parallel to PyVBMC, a new software for sample-efficient Bayesian inference \parencite{huggins2023pyvbmc, acerbi2018variational, acerbi2020variational, acerbi2019exploration}. PyBADS can be used in combination with PyVBMC, by providing an effective way of initializing the inference algorithm at the maximum-a-posteriori (MAP) solution.

\hypertarget{applications-and-usage}{%
\subsection{Applications and usage}\label{applications-and-usage}}

The BADS algorithm, in its MATLAB implementation, has already been applied in multiple fields, especially in neuroscience where it finds a broad audience by efficiently solving difficult model-fitting problems \parencite{cao2019causal, Tajima2019, Li2020, DAUBE20191924}. Other fields in which BADS has been successfully applied include control engineering \parencite{stenger2022benchmark}, electrical engineering \parencite{li2022topology}, material engineering \parencite{ren2021novel}, robotics \parencite{REN2020575}, petroleum science \parencite{FENG20222879}, environmental economics \parencite{Wildfires2020}, and cognitive science \parencite{STENGARD2022105160, vanOpheusden2023}. Moreover, BADS has been shown to perform best in most settings of a black-box optimization benchmark for control engineering \parencite{stenger2022benchmark}, highlighting the effectiveness of our algorithm compared to other BayesOpt and direct-search approaches.
With PyBADS, we bring the same sample-efficient and robust optimization to the wider open-source Python community, while improving the interface, test coverage, and documentation.

The package is available on both PyPI (\texttt{pip\ install\ pybads}) and \texttt{conda-forge}, and provides an idiomatic and accessible interface, only depending on standard, widely available scientific Python packages \cite{harris_array_2020}. The user only needs to give a few basic details about the objective function and its parameter space, and PyBADS handles the rest of the optimization task. PyBADS includes automatic handling of bounded variables, robust termination conditions, sensible default settings, and does not need tunable parameters. At the same time, experienced users can easily supply their own options. We have extensively tested the algorithm and implementation details for correctness and performance. We provide detailed \href{https://github.com/acerbilab/pybads/tree/main/examples}{tutorial}, so that PyBADS may be accessible to those not already familiar with black-box optimization, and our comprehensive \href{https://acerbilab.github.io/pybads}{documentation} will aid not only new users but future contributors as well.

\hypertarget{acknowledgments}{%
\section{Acknowledgments}\label{acknowledgments}}

We thank Bobby Huggins, Chengkun Li, Marlon Tobaben and Mikko Aarnos for helpful comments and feedback.
Work on the PyBADS package is supported by the Academy of Finland Flagship programme: Finnish Center for Artificial Intelligence FCAI.

\newrefcontext[sorting=nyt]
\printbibliography

\end{document}